# Exploring Values in Museum Artifacts in the SPICE project: a Preliminary Study

Nele Kadastik*
neka@create.aau.dk
Aalborg University
Aalborg, Denmark

Thomas A. Pedersen*
tape@create.aau.dk
Aalborg University
Aalborg, Denmark

Luis Bruni*
leb@create.aau.dk
Aalborg University
Aalborg, Denmark

Rossana Damiano*
rossana.damiano@unito.it
Università di Torino
Torino, Italy

Antonio Lieto*
antonio.lieto@unito.it
Università di Torino
Torino, Italy

Manuel Striani*
manuel.striani@unito.it
Università di Torino
Torino, Italy

Tsvi Kuflik*
tsvikak@is.haifa.ac.il
University of Haifa
Haifa, Israel

Alan Wecker*
ajwecker@gmail.com
University of Haifa
Haifa, Israel

## ABSTRACT

This document describes the rationale, the implementation and a preliminary evaluation of a semantic reasoning tool developed in the EU H2020 SPICE project to enhance the diversity of perspectives experienced by museum visitors. The tool, called DEGARI 2.0 for values, relies on the commonsense reasoning framework $T^{CL}$, and exploits an ontological model formalizing the Haidt's theory of moral values to associate museum items with combined values and emotions. Within a museum exhibition, this tool can suggest cultural items that are associated not only with the values of already experienced or preferred objects, but also with novel items with different value stances, opening the visit experience to more inclusive interpretations of cultural content. The system has been preliminarily tested, in the context of the SPICE project, on the collection of the Hecht Museum of Haifa.

## 1 INTRODUCTION

It is recognized that culture and cultural heritage, by blending past, present and future, play a vital role in driving more cohesive and inclusive societies, and in their essence imply a shared sense of belonging and purpose [24]. While the importance of cultural participation has been widely acknowledged and demonstrated – particularly for its "potential to tackle exclusion" [2] – more re-search is needed to better understand the relations and underlying reciprocal dynamics and processes that could conceivably support it. Cultural institutions and museums play a key role in mediating and assisting citizens in participating in the cultural process. Thus, to ensure culture as a platform for sociability and for reinforcing belonging and identity, cultural heritage institutions should be seen as a key enabler. When cultural heritage and art can be seen as the core around which new experiences, relationships and knowledge is built, heritage institutions hold the possibility for effectively facilitating it. Murzyn-Kupisz and Dziazek [25] propose that heritage institutions function as 'community hubs', or in other words "spaces where trust is built and social networks are created". It is therefore more important than ever that heritage institutions actively seek new innovative ways to engage diverse audiences in participating in cultural heritage.

SPICE (Social Cohesion, Participation, and Inclusion through Cultural Engagement) is an EU H2020 project, dedicated to the development of new methods and digital tools for enhancing cultural experiences in museums through citizen curation of cultural heritage. Citizen curation in SPICE offers a highly personalised mode of participation in which citizens apply "curatorial meth-ods to archival materials available in memory institutions in order to develop their own interpretations, share their own perspective and appreciate the perspectives of others" [1]. The repertoire of activities offered by the platform enables different combinations of hybrid participation in the pre-, during and post- visit. The SPICE project is working with co-designing, and testing with five museum partners, in five different countries, involving diverse communities of citizens, where a particular focus is given to engaging different minority groups that tend to be under-represented in cultural activities[1].

The underlying objectives of SPICE can be considered twofold. Firstly, it is anticipated that the interpretive and reflective processes embedded in SPICE will unveil new personalised perspectives about the perceived artefacts (and their histories). Secondly, by drawing on similarities, differences and relations among citizens' interpretations of cultural heritage, whatever code they use, it is expected that these processes will be able to tell something about the citizens themselves. The tools and methods developed in SPICE can be used to support "reflection within and across groups", to highlight emerging communities to which the citizens can, explicitly or implicitly, recognize themselves as belonging to. Thus, an important goal in SPICE is to develop new technologies and tools that can support citizens in contributing rich interpretations, but also in sharing reflections on the interpretations of others. In this paper we describe the design logic, the implementation and a preliminary evaluation of one of the semantic reasoning tools being integrated in the SPICE platform [5].

The core reasoning tool, called DEGARI 2.0, relies on a probabilistic extension of a typicality-based Description Logic called $T^{CL}$, (Typicality-based Compositional Logic, introduced in [18]). This framework allows one to describe and reason upon an ontology with commonsense (i.e. prototypical) descriptions of value concepts, as well as to dynamically generate novel prototypical concepts in a knowledge base as the result of a human-like recombination of the existing ones. One of the advantages of the this underlying logic-based approach lies in the possibility of providing explicit explanations of the values associated to each recommendation [20], in line with the requirements posed by the notion of Trustworthy Artificial Intelligence[2].

## 2 EMOTIONS AND VALUES IN THE SPICE FRAMEWORK

As all cultural heritage rests on a complex interlinked system of meaning, expression of values, attitudes, beliefs, knowledge, skills and traditions that continue to transform and evolve in time [21], SPICE is developing and integrating a suite of reasoning tools (including ontologies, semantic reasoners, a Linked Data hub, and content recommender), to personalise the visitor experience based on different attributes such as, values, emotions, interests and themes. Hence, for the museum users, DEGARI 2.0 plays a major role in classifying and suggesting cultural items, which are associated not only with the very same value of already experienced or preferred objects (e.g. within a museum exhibition), but also novel items that share different values stances. The logic of DEGARI 2.0 builds on a previously developed logic for deriving emotions with the system of DEGARI [3, 20], and expands upon this by combining emotions and moral values. As such, the following section discusses these two concepts, and their ties to one another.

Emotions can be regarded as a principal part of consciousness, as "[b]y virtue of being born, the person has the ability to experience pleasant feelings, or positive affect, and unpleasant feelings, or negative affect" [22]. Hence, emotions are considered to permeate all human experience, and as such, any aesthetic experience [20]. As cultural objects and historical events often suggest an emotional bond, emotions can be argued to be closely tied to engagement with culture and cultural heritage. Within SPICE, we rely precisely on this relationship, as it can potentially result in citizens becoming more inclined to contribute "rich" narratives. In this sense, cultural heritage can revive dormant emotions, and potentially serve as an incentive for personal storytelling.

---

[1] https://spice-h2020.eu/
[2] https://digital-strategy.ec.europa.eu/en/library/ethics-guidelines-trustworthy-ai

While the definition of emotions, and its ties to other psychological concepts such as cognition, motivations and values, is an ongoing discussion in the field of affective psychology [17, 27], in SPICE, we have adopted the conceptualization of emotions by Plutchik [27]. Similarly to the highly influential conceptualization of basic emotions by Paul Ekman [6], Plutchik considered a range of basic emotions (eight to be specific). However, Plutchik further expanded upon this, by taking into account varying intensities, as well as different combinations of the basic emotions. Plutchik considered these eight basic emotions, and their compound derivatives, as adjacent and opposing emotions, organized in the "Plutchik Wheel of Emotions" [28]. It is also this model that has been integrated in DEGARI, Dynamic Emotion Generator and Reclassifier [20], [19] which computes the compound emotions from information regarding the basic emotions derived through natural language processing. With this outset, we propose to derive values based on their emotional properties.

The concept of values is widely accepted as a key component of personality, or identity [4, 13, 15, 22, 23, 29, 31, 32, 34–36]. However, as opposed to emotions that are considered a genetic, inborn ability of human conception [22], values are regarded as being "determined by the nature of the culture in which the person exists" and are thus "instilled in you by society" ([22] p. 197, 210). Generally, values are agreed to be represented as desirable behaviour and end-states, and having the characteristic of being transsituational [13, 26, 31, 34, 35]. While a number of conceptualizations of values exist, the following will consider the concept of moral values, which serves as the basis for DEGARI 2.0. Rokeach [31] regarded the concept of moral values as "narrower than the general concept of values", as moral values were considered to mainly regard desirable behaviour, and "arouse pangs of conscience or feelings of guilt for wrongdoing". As such, moral values can be viewed as relating to "social motivations beyond direct self-interest" ([12] p. 998). An influential conceptualization of moral values has been introduced by Jonathan Haidt and colleagues in their Moral Foundations Theory (MFT) [10, 12]. A central notion of MFT, is that multiple moral foundations underlie morality. As such, the theory expands beyond aspects of harm and fairness, traditionally emphasised by moral psychology [9, 12, 16, 37]. The basic model of the MFT initially suggested five foundations for morality, represented as: (1) Care/Harm, (2) Fairness/Cheating, (3) Loyalty/Betrayal, (4) Authority/Subversion, and (5) Sanctity/Degradation. However, both its founders and supporters suggest that more foundations may exist. With MFT, Haidt additionally diverges from a common focus on moral reasoning as the primary driver for moral judgement. Instead, he highlights moral intuitions (or moral emotions) as the driving factor for moral judgement, with moral reasoning occurring as an ex post facto response. In other words, "[o]ne sees or hears about a social event and one instantly feels approval or disapproval" ([11] p. 818). As such, moral emotions are considered as conscious, but sudden, responses in the form of affective valence towards something. Hence, moral intuitions serve moral judgement before moral reasoning is employed to provide the rational arguments for the sensations felt [9, 13]. This close connection between the moral intuitions and emotions can be seen as a unique feature of the model,and is key to its utilisation in the DEGARI 2.0 pipeline for building Hybrid Moral Values.

## 3 HYBRID MORAL VALUES IN SPICE

The logic $T^{CL}$, that we recall here for self-containedness, is the result of the integration of two main features: (i) an extension of a non-monotonic Description Logic of typicality ALC +$T_R$ introduced in [7, 8] with a distributed semantics; (ii) a well established heuristics inspired by cognitive semantics for concept combination and generation ([14]), in order to formalize a dominance effect between the concepts to be combined: for every combination, it distinguishes a HEAD, representing the stronger element of the combination, and a MODIFIER. The basic idea is to extend an initial knowledge base (ontology) with a prototypical description of a novel concept, obtained by the combination of two existing ones, namely a HEAD concept and a MODIFIER concept. In this logic, typical properties can be directly specified by means of a typicality operator T enriching the underlying Description Logic, and a knowledge base can contain inclusions of the form $p :: T(C) \sqsubseteq D$ to represent that "typical Cs are also Ds", where p is a real number between 0.5 and 1, representing the probability of finding elements of C being also D. From a semantic point of view, it considers models equipped by a preference relation among domain elements, where $x < y$ means that x is "more normal" than y, and that the typical members of a concept C are the minimal elements of C with respect to this relation. An element x is a typical instance of a given concept C if x belong to the extension of the concept C, written $x \in C^I$, and there is no element in $C^I$ "more normal" than x. $T^{CL}$ also considers the key notion of scenario. Intuitively, a scenario is a knowledge base obtained by considering all rigid properties as well as all ABox facts, but only a subset of typicality properties. To this aim, it considers an extension of the Description Logic ALC + $T_R$ based on the distribution semantics known as DISPONTE ([30]). The idea is to assume that each typicality inclusion is independent from each other in order to define a probability distribution over scenarios: roughly speaking, a scenario is obtained by choosing, for each typicality inclusion, whether it is considered as true of false. Reasoning can then be restricted to either all or some scenarios. $T^{CL}$ equips each scenario with a probability, easily obtained as the product, for each typicality inclusion, of the probability p in case the inclusion is involved, $(1 - p)$ otherwise. It immediately follows that the probability of a scenario introduces a probability distribution over scenarios, that is to say the sum of the probabilities of all scenarios is 1.

In the context of our system, $T^{CL}$ allows us to provide a formal, explainable framework for combining prototypical descriptions of value concepts. More in detail: $T^{CL}$ is used to create hybrid value concepts obtained by combining emotional features extracted from lexical resources (since typically values are associated with emotional nuances) with the specific values features extracted for the values in hand. In particular, the information about the emotional concepts and their corresponding features to combine via $T^{CL}$ are extracted from the NRC Emotion Intensity Lexicon ([33][3]). This lexicon associates, in descending order of frequency, words to emotional concepts and, for our purposes, we considered the most frequent terms available in such lexicon as typical features of the basic emotions. For the characterization of the typical features constituting the Value classes we used the eMFD [38], extended Moral Foundation Dictionary, assigning probability to each value concept. The idea is to create hybrid typical value/emotion concepts combining eMFD vocabulary and the emotional prototypes already used in [20]. Table 1 shows the manual mapping provided between Haidt model and Plutchik: moral emotions (from Haidt's model) are mapped onto the corresponding values (MFT), then onto Plutchik's emotions.

The obtained synthetic "symbolic hybrids" (they only contain up to 7 features for each generated concepts) are prototypes of value concepts able to capture in a better way the strong connection between emotions and values. Overall, once the association of lexical features to the emotional and value concepts in the is obtained and the hybrid emotion-values concepts are generated via the logic $T^{CL}$, the system is able to reclassify cultural items (described in some catalogue), or textual descriptions associated to that cultural items.

The current version of the system, available as a web service, accepts JSON files containing a textual description of the cultural items and performs an automatic information extraction step generating a lemmatized version of the JSON descriptions of the cultural item and a frequentist-based extraction of the typical terms associated to each cultural item in its textual description (the assumption is that the most frequently used terms to describe an item are also the ones that are more typically associated to it). The frequencies are computed as the proportion of each term with respect to the set of all terms characterizing the item, in order to compare. These two tasks are performed by using standard libraries like Natural Language Toolkit[4] and TreeTagger[5]. Once this pre-processing step is automatically done, the final representation of the cultural items is compared with the representations of the typical compound values obtained with $T^{CL}$: if a cultural item contains all the rigid properties and at least the 30% of the typical properties of the compound emotion under consideration, then the item is classified as belonging to it.

As anticipated before, the value model relies on the moral foundations theory. In particular, it relies on the five foundations for which the evidence is best:

- Care/harm: related to our evolution as mammals with attachment systems and an ability to feel (and dislike) the pain of others, it underlies virtues of kindness, gentleness, and nurturance.

---

[3] Such lexicon provides a list of English words, each with real-values representing intensity scores for the eight basic emotions of Plutchik's theory. The intensity scores were obtained via crowdsourcing, using best-worst scaling annotation scheme.
[4] https://www.nltk.org/
[5] https://www.cis.uni-muenchen.de/ schmid/tools/TreeTagger/

Table 1: Manual mapping between Haidt's model of moral values theory and Plutchik's theory

| Emotion | Value | Mapped emotion |
|---|---|---|
| Admiration | Authority | Awe |
| Anger | Cheating (Fairness -) | Anger |
| Compassion | Harm (Care -) | Grief, Sadness, Pensiveness |
| Contempt | Betrayal (Loyalty -), Cheating (Fairness -) | Disapproval |
| Disgust | Degradation (Sanctity -) | Disgust, Loathing |
| Embarrassment | Cheating (Fairness -) | Annoyance |
| Evaluation | Sanctity | Awe |
| Fear | Subversion (Authority -) | Terror |
| Gratitude | Fairness | Vigilance, Anticipation, Interest |
| Guilt | Cheating (Fairness -) | Remorse |
| Pity | Harm (Care -) | Grief, Sadness, Pensiveness |
| Pride | Loyalty | Admiration, Trust, Acceptance |
| Rage | Betrayal (Loyalty -) | Rage |
| Remorse | Harm (Care -) | Grief, Sadness |
| Reproach | Betrayal (Loyalty -) | Aggressiveness |
| Respect | Authority | Submission, Fear |
| Shame | Loyalty - | Remorse |

- Fairness/cheating: related to the evolutionary process of reciprocal altruism, it generates ideas of justice, rights, and autonomy.
- Loyalty/betrayal: related to our history as tribal creatures able to form shifting coalitions, it underlies virtues of patriotism and self-sacrifice for the group. It is active anytime people feel that it's "one for all, and all for one".
- Authority/subversion: shaped by our primate history of hierarchical social interactions, it underlies virtues of leadership and followership, including deference to legitimate authority and respect for traditions.
- Sanctity/degradation: shaped by the psychology of disgust and contamination, it underlies religious notions of striving to live in an elevated, less carnal, more noble way, and the widespread idea that the body is a temple which can be desecrated by immoral activities and contaminants.

## 4 EVALUATION

In order to test the application of value detection and value rea-soning in relation with museum exhibits, we have preliminarily evaluated our system on data coming from Hecht Museum at the University of Haifa (Israel). The system has been launched on an experimental subset of 12 items, selected by the museum curators to stimulate diverse perspectives. For the description of the items, the relevant pages of the English edition of Wikipedia were considered.

Figure 1 shows the classification process for the item "Catapult" (on the left of the figure). It is possible to see that DEGARI 2.0 takes into account the prototypes (synthetic "symbolic hybrids") {degradation} and {disgust} (middle column in the figure) and re-combines them into another concept called {degradation-disgust} for classifying the item (left column). Notice that the relevant key-words for the association with the prototypes are reported in bold.

Table 2 shows the results for DEGARI 2.0 values classification on three Hecht Museum's items, based on the corresponding Wikipedia pages. In particular, the item "Catapult" is classified by DEGARI with the {degradation-disgust} combined concept (column DE-GARI 2.0 with values) because the description contains the key-words "molestation" and "weapons" (column words matches) that trigger the classification. The moral value 'degradation' (mapped as "Sanctity⁻" in MFT) is triggered by the keyword "weapon". In addition to generating the association with values through the map-ping shown in Table 1 (which shows the manual mapping provided between values and emotions), the system also generates the association with the emotions; for "Catapult", (Table 2), the keyword "molestation" acts as a trigger for the emotion "disgust" (column emotions), associated with that particular item. The item "Roman war gear", has been classified as {betrayal-aggressiveness} because the keywords "brutality" and "violently", contained in its descrip-tion, trigger the emotion "aggressiveness", which is mapped onto "betrayal" ("Loyalty⁻" in MFT, as illustrated in Table 1). The item "Bar Kochva Rebellion" (last row), has been classified as {sanctity-awe}, because the keywords "surprise", "torture" and "kill" trigger the emotion "awe", which is mapped onto "Sanctity.

## 5 CONCLUSION AND FUTURE WORK

In this paper, we discussed how cultural heritage institutions, by finding new innovative ways to engage their audiences, can re-inforce belonging and identity. In the context of the EU Horizon 2020 project SPICE, which focuses on the development of new methods and digital tools for citizen curation of cultural heritage, we presented a novel reasoning system which classifies museum items with value-emotion associations. For this experiment, only the relevant Wikipedia pages were considered, resulting in a limited number of classifications. In the future, however, the texts attached to the items are expected to include the comments and annotations provided by curators and citizens, thus opening the way to more classifications. The mapping of the items in the museum collection onto the combined values and emotions opens novel perspectives for the visitors interacting with the items. On the one side, it allows them to locate themselves in the value-emotion system, raising their awareness on multiple points of views. On the other side, it enables the recommendation of similar and opposite items based on values and emotions, which in turn enable forms of critical ex-ploration of cultural items and (self)reflection on identity, bringing visitors out of the so-called echo chambers.

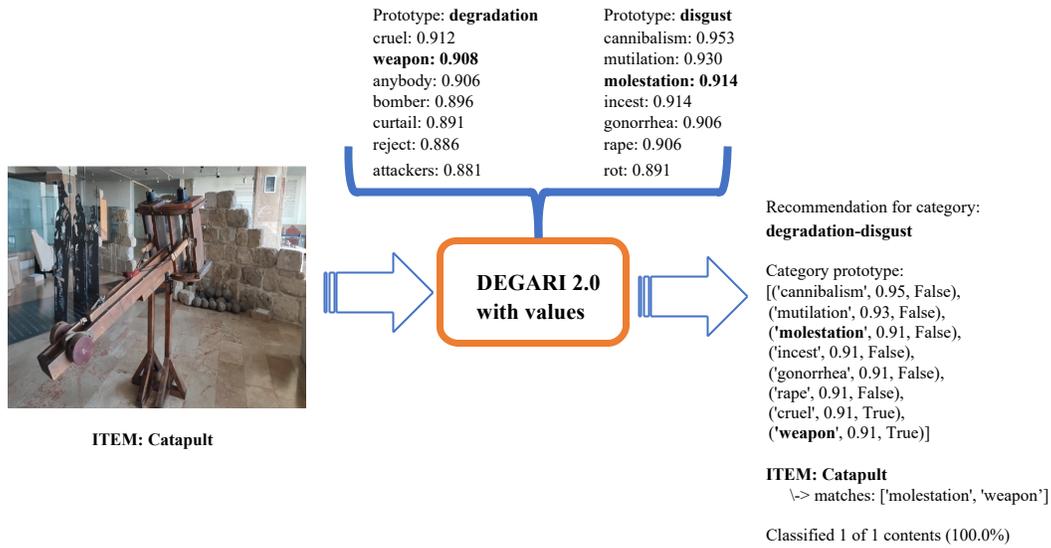

Figure 1: Example of DEGARI 2.0 moral value classification for item "Catapult". The item has an associated textual description retrieved from Wikipedia.

Table 2: DEGARI 2.0 moral values classification with Hecht dataset

| Item | DEGARI 2.0 with values | words matches | emotions | | | moral values | | |
|---|---|---|---|---|---|---|---|---|
| | | | disgust | aggressiveness | awe | degradation (Sanctity -) | betrayal (Loyalty -) | sanctity |
| Catapult | degradation-disgust | ['molestation', 'weapon'] | molestation | | | weapon | | |
| Roman war gear | betrayal-aggressiveness | ['brutality', 'violently'] | | brutality, violently | | | | |
| Bar Kochva Rebellion | sanctity-awe | ['surprise', 'torture', 'kill'] | | | surprise, torture, kill | | | |

## ACKNOWLEDGMENTS

The research leading to these results/this publication has been par-tially funded by the European Union's Horizon 2020 research and innovation programme http://dx.doi.org/10.13039/501100007601 un-der grant agreement SPICE 870811. The publication reflects the author's views. The Research Executive Agency (REA) is not liable for any use that may be made of the information contained therein.